# A Time-to-first-spike Coding and Conversion Aware Training for Energy-Efficient Deep Spiking Neural Network Processor Design

Dongwoo Lew, Kyungchul Lee, and Jongsun Park
School of Electrical Engineering, Korea University, Seoul, Korea
{wwe9712, 1225lkc, jongsun}@korea.ac.kr

## ABSTRACT

In this paper, we present an energy-efficient SNN architecture, which can seamlessly run deep spiking neural networks (SNNs) with improved accuracy. First, we propose a conversion aware training (CAT) to reduce ANN-to-SNN conversion loss without hardware implementation overhead. In the proposed CAT, the activation function developed for simulating SNN during ANN training, is efficiently exploited to reduce the data representation error after conversion. Based on the CAT technique, we also present a time-to-first-spike coding that allows lightweight logarithmic computation by utilizing spike time information. The SNN processor design that supports the proposed techniques has been implemented using 28nm CMOS process. The processor achieves the top-1 accuracies of 91.7%, 67.9% and 57.4% with inference energy of 486.7uJ, 503.6uJ, and 1426uJ to process CIFAR-10, CIFAR-100, and Tiny-ImageNet, respectively, when running VGG-16 with 5bit logarithmic weights.

## 1 INTRODUCTION

Inspired by the low power nature of the human brain, spiking neural networks (SNNs) mimic the behaviors of neurons and synapses of human's neuro-biological systems. By processing information using binary spikes in an event-driven fashion, SNN is expected to be implemented with an energy-efficient way. However, compared to the artificial neural networks (ANNs) that shows remarkable performance in a wide variety of applications, SNN still suffers from relatively low recognition accuracies [1]. In order to address the accuracy issues, many previous research works have been focused on improving the performance of SNN training. In [2], surrogate gradients are utilized, and bio inspired algorithm [3] is also used for direct training of SNN. But, those approaches suffer from still low accuracies compared to ANN. ANN to SNN conversions [4], [5] recently achieve the accuracies comparable to ANNs. In ANN to SNN conversion, ANN is first trained using standard backpropagation, then it is converted to SNN by applying various conversion techniques like weight normalization [5] and gradient-based optimization [4]. Although considerable accuracy improvement has been achieved with the algorithmic advances, the improved accuracy comes at the expense of deeper and larger networks that leads to large computational complexities to process the SNNs [6].

In addition to the algorithmic advances, running large and deep SNNs on hardware in an energy-efficient way is another important issue to consider. A simple approach would be employing the conventional graphic processing units (GPUs) or ANN accelerators to process the SNN operations. However, due to irregular sparsity caused by event-driven property and timesteps of SNN, running SNNs on GPU or ANN accelerator can incur considerable throughput and energy efficiency degradations [1], [7]. To seamlessly process the sparse SNN operations that include irregular and repetitive memory accesses, dedicated SNN processors have been designed from industry and academia. Those processors includes IBM TrueNorth [8], Intel Loihi [9], Tianjic [10], SpinalFlow [7], and FlexLearn [11]. Although the dedicated SNN processors improve energy efficiency as well as throughput based on the efficient dataflows and the architectures suited for SNNs, the hardware efficiencies and the accuracies of the SNN processors still fall behind those of ANNs.

In this paper, we propose an SNN processor, which can efficiently process deep SNN with much improved accuracy. As a basic architecture, we modify the SpinalFlow [7] and improve the design to be able to process the existing state-of-the-art temporal SNN models. First, to reduce the ANN-to-SNN conversion loss, we propose a conversion aware training (CAT). In the proposed CAT, by modifying the activation functions, the forward propagation of SNN is simulated during ANN training to reduce the conversion error. Second, to reduce the overall computation complexity for processing spikes, a time-to-first-spike (TTFS) coding that allows lightweight logarithmic computation, is presented. The SNN processor that can run the SNN model obtained by the proposed training, has been implemented using 28nm CMOS process. The hardware implementation results show that the SNN processor can run larger datasets, and it can compete with previous ANN processors in terms of throughput and energy consumption.

The rest of paper is organized as follows. In Section 2, the preliminary of SNNs and the time-to-first spike neural network (T2FSNN), which is the baseline TTFS coding, are presented. The conversion aware training and TTFS coding that allow logarithmic computing are proposed in Section 3. The hardware architecture supporting the proposed approach and implementation results are presented in Section 4 and Section 5, respectively. Finally, conclusions are drawn in Section 6.

## 2 BACKGROUND

### 2.1 Spiking Neural Network

SNNs transfer data between neurons using discrete spikes that are spread out in the time domain. Here, the output spikes of neuron $i$ in layer $l$ are described as

$$S_i^l(t) = \sum_{t_i^{l,(f)} \in F_i^l} \delta(t - t_i^{l,(f)}), \quad (1)$$

where $\delta(t)$ is the Dirac delta function, $f$ is the index of a spike, and $F_i^l$ is a set of spikes at time $t$ meeting the fire condition explained as

$$t_i^{l,(f)} : u_i^l(t_i^{l,(f)}) \geq \theta_i^l(t_i^{l,(f)}), \quad (2)$$

where $u_i^l(t)$ is the membrane voltage of neuron and $\theta_i^l(t)$ is the threshold voltage at time $t$. These output spikes are sent to the neurons in the next layer and get integrated into the membrane potential of the neurons. Among many neuron types, integrate and fire (IF) neuron is widely used because of its simplicity. Its operation when input spikes get integrated is explained as

$$u_j^l(t) = u_j^l(t-1) + z_j^l(t), \quad (3)$$

where $z_j^l$ is the sum of postsynaptic potential (PSP), which can be described as

$$z_j^l(t) = \sum_i w_{ij}^l d_j^l(t) S_i^{l-1}(t) + b_j^l, \quad (4)$$

where $w_{ij}^l$ is the synaptic weight, $d_j^l$ is the dendrite function, and $b_j^l$ is a bias. Because neurons emit spikes only when (2) is met, spikes are sparse and irregular in the time domain.

### 2.2 T2FSNN: Kernel-based TTFS Coding

T2FSNN [4] is a kind of TTFS coding with at most one spike per neuron for ANN-to-SNN conversion. The overview of T2FSNN is presented in figure 2. To encode the data of a neuron in a single spike, an efficient temporal coding is realized using the kernel-based dynamic threshold and the dendrite. As shown in figure 2, first, an IF neuron of T2FSNN has two separate phases, fire (encoding) phase and integration (decoding) phase. Neuron enters the fire phase after the end of integration phase. During each phase, a neuron encodes or decodes spikes using the threshold kernel or the dendrite kernel for a given time window $T$. The kernels decrease monotonically as

$$\epsilon^l(t - t_{ref}^l) = \exp(-(t - t_{ref}^l - t_d^l)/\tau^l), \quad (5)$$

where $t_{ref}^l$ is the start time of fire phase, $t_d^l$ and $\tau^l$ are the delay time and the time constant of each layer, respectively. Using this kernel, pre-synaptic neurons in the fire phase emit spike (encode) when its membrane voltage exceeds the threshold. The threshold is explained as

$$\theta^l(t) = \theta_0 \epsilon_{FI}^l(t - t_{ref}^l), \quad (6)$$

where $\theta_0$ is the base threshold and $\epsilon_{FI}^l$ is the fire kernel. As the threshold exponentially decreases as time passes, the neurons that have larger membrane voltage fire spike earlier. The encoded spikes are propagated to post-synaptic neurons in the next layer,

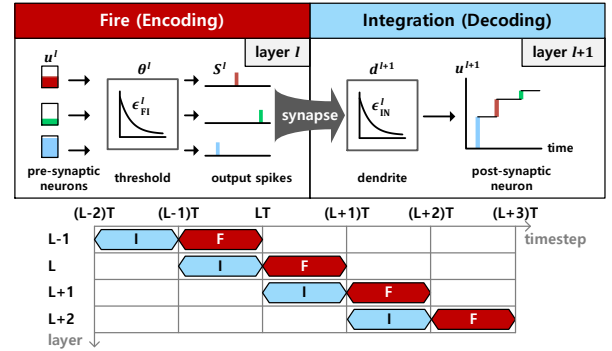

Figure 1: The overview of T2FSNN.

and the neurons in the integration phase decode input spikes to PSP using the following dendrite equation

$$z_j^l(t) = \sum_i w_{ij}^l \epsilon_{IN}^l(t^{l-1} - t_{ref}^{l-1}) + b_j^l, \quad (7)$$

where $\epsilon_{IN}^l$ is the integration kernel. In T2FSNN, $\tau^l$ and $t_d^l$ of the integration kernel set to be equal to the fire kernel of the previous layer to make decoding accurate.

## 3 CONVERSION AWARE TRAING AND TTFS CODING FOR LOGARITHMIC COMPUTATION

This section first proposes a hardware-friendly training method for kernel-based TTFS coding that efficiently reduces the ANN-to-SNN conversion error by adjusting the activation function during training. Based on the training method, the constraints of the kernel-based TTFS coding that enables hardware-friendly logarithmic computation is decided, which can be used in conjunction with the logarithmic quantization of weights.

### 3.1 Proposed Conversion Aware Training (CAT)

When an ANN is converted to SNN, an inevitable accuracy loss occurs due to the error induced by the data representation change from analog values to discrete spikes. Reducing this error is one of the keys to successful ANN-to-SNN conversion. In T2FSNN, the conversion error reduction is performed by tuning $t_d^l$ and $\tau^l$ using post-conversion optimization technique [4], where the error introduced during spike encoding and decoding process of each layer is formulated in each layer. Then, by tuning $t_d^l$ and $\tau^l$ (using the gradient descent algorithm), the layer-wise coding error is reduced, thus decreasing the conversion loss. Although this conversion error reduction approach achieves improved accuracy in SNN, but it makes the SNN model have various TTFS kernels for every layer due to different $t_d^l$ and $\tau^l$. When the SNN model is implemented in hardware, it incurs excessive hardware burden of requiring multiple (or reconfigurable) spike encoding and decoding units.

In this work, we propose a pre-conversion optimization approach of conversion aware training (CAT) to reduce ANN-to-SNN conversion loss without employing tunable parameters in SNN. Here, the basic idea is that we can intentionally make the ANN to learn the data representation of SNN while ANN training. This can be done during ANN training by simulating SNN forward propagation using only the permitted values of SNN data representation method (e.g. TTFS coding) to propagate the ANN activation. In order to simulate SNN forward propagation, the first thing to do is finding an activation function that can simulate TTFS coding during ANN training. So, we first examine the timestep of spike $t^l$ obtained by combining (2) and (6) as following:

$$t^l = \lceil \tau^l \ln(u_i^l(t_{ref}^l - 1)/\theta_0) + t_d^l \rceil + t_{ref}^l . \quad (8)$$

While (8) is not directly calculated during runtime in hardware, we can use (8) to show that $t^l$ can be interpreted as the quantized form of $u_i^l$ due to the ceiling operation in (8). As we consider that reducing quantization error is the similar process with reducing conversion error, we come up with an idea that modifying activation functions, which is widely used in quantization aware training to simulate quantization, can be efficiently adopted to reduce the conversion error. So, we design an activation function that simulates TTFS coding during ANN training.

Then, we first define a new TTFS kernel based on base 2 as

$$\kappa^l(t - t_{ref}^l) = 2^{-(t-t_{ref}^l)/\tau} . \quad (9)$$

Compared to the original kernel $\epsilon^l$, the new kernel $\kappa^l$ does not have $t_d^l$, and $\tau$ is no longer a layer-wise parameter, but a single parameter shared by all the layers. The new base 2 kernel is also designed for enabling logarithmic computation in hardware, which will be discussed in the next section. If appropriate $T$ and $\tau$ is chosen and exponential identity is used to convert base of kernel, using the new kernel does not directly affect classification accuracy since $\kappa^l$ is almost identical to the original kernel $\epsilon^l$.

Based on $\kappa^l$, the following activation and derivative are derived and used during conversion aware training (CAT)

$$\phi_{TTFS}(x) = \begin{cases} 0, & x < \kappa^l(T - t_{ref}^l) \\ 2^{\lceil \tau \log_2(x/\theta_0) \rceil}, & \kappa^l(T - t_{ref}^l) \leq x < \theta_0 \\ \theta_0, & otherwise \end{cases}, \quad (10)$$

$$\frac{d\phi_{TTFS}}{dx} = \begin{cases} 1, & \kappa^l(T - t_{ref}^l) \leq x < \theta_0 \\ x, & otherwise \end{cases}, \quad (11)$$

where $\phi_{TTFS}$ is an activation that simulates the TTFS encoding and decoding. Figure 2 shows the activation functions and the data representation errors in the conversion with varying inputs. As shown in the figure, TTFS activation ($\phi_{TTFS}$) has no error when compared to the coding used in SNN. This ensures minimal conversion loss and also eliminates the need of weight normalization [5] after conversion (with exception of the output layer, as there is no activation function).

Although $\phi_{TTFS}$ well simulates the SNN after conversion, stability issue can be introduced as $\phi_{TTFS}$ is a discrete function. In this work, to address the stability issue, a relaxed activation (called clip) is adopted during the early stage of the training to allow the network to enter stable state. This relaxed activation is expressed as

$$\phi_{Clip}(x) = \text{clip}(x, \theta_0, 0), \text{ where} \quad (12)$$

$$\text{clip}(x, max, min) = \begin{cases} max, & max \leq x \\ x, & min < x < max \\ min, & otherwise \end{cases}. \quad (13)$$

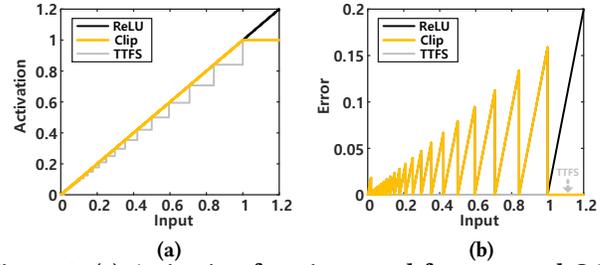

Figure 2: (a) Activation functions used for proposed CAT and (b) error compared to SNN ($T = 24, \tau = 4, \theta_0 = 1$)

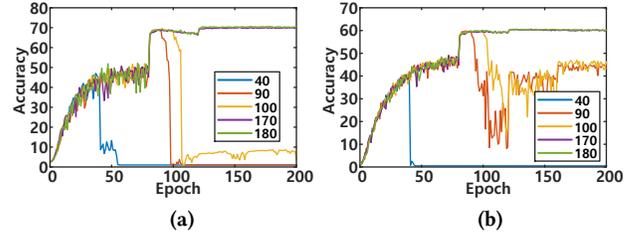

Figure 3: VGG-16 (ANN) test accuracy during training with different $\phi_{TTFS}$ applying epoch (a) cifar-100, (b) Tiny-ImageNet ($T = 24, \tau = 4, \theta_0 = 1$)

When the relaxed activation is used, it shows data representation error as shown in Figure 2(b) (clip in the figure), but ensures stable training.

Before discussing the whole training procedure, we need to consider the kinds of the activation functions to be used and the applying orders of activation functions, based on the purpose and the characteristic of each activation function. While $\phi_{Clip}$ and $\phi_{TTFS}$ should be sufficient for simulating SNN during ANN training, addition of ReLU can be helpful during initial stage of training, which is widely used in QAT as well [12]. So, we use ReLU, $\phi_{Clip}$ and $\phi_{TTFS}$ in our training. Regarding the order of activation function used, ReLU is initially used to boost initial training, then $\phi_{Clip}$ is used during the bulk of training in order to train ANN while maintaining stability at the cost of slight error in SNN simulation. Lastly, $\phi_{TTFS}$ is applied after $\phi_{Clip}$ to allow ANN to get trained with accurate simulation of SNN.

As we use multiple activation functions in training, the specific time to switch the activations should be decided through the simulations. In our simulations, we trained VGG-16 using stochastic gradient descent (SGD) with the momentum of 0.9 and the weight decay of 5E-4 for 200 epochs and adopt the batch normalization. Learning rate starts at 0.1 and get divided by 10 on 80, 120, and 160 epochs.

In our ANN training procedure, the network is trained using ReLU for 10 epochs at the start of training, then the activation function is switched to $\phi_{Clip}$ and trained. Now, we need to find the starting epoch of applying $\phi_{TTFS}$ to the network. In our simulations, we select and test various epochs to switch from $\phi_{Clip}$ to $\phi_{TTFS}$ based on the learning rate, and figure 3 shows the test accuracies obtained from our simulations. As shown in figure 3, starting to apply $\phi_{TTFS}$ at the epochs earlier than 159 when the learning rate is larger than 1E-3, crashes the training, while applying $\phi_{TTFS}$ after epoch 160 while learning rate is 1E-4 shows stable training. Based on the observations, epoch 170 is selected for applying $\phi_{TTFS}$ to the network. After the activation is switched to $\phi_{TTFS}$ and the training continues until epoch 200. In the training process, $\phi_{TTFS}$ is appended to the input of the first hidden layer at the first epoch to simulate input image being presented using spikes. After the ANN is trained, ANN-to-SNN conversion is performed to obtain SNN model. During conversion, batch normalization layers are fused into the weights of convolution layers and the weight normalization is applied to the output layer [5].

In the CAT procedure described above, there are multiple components that partially contribute the overall conversion loss reduction. To analyze the effects of individual components of CAT on the classification accuracy and conversion loss ($acc_{SNN} - acc_{ANN}$), simulations are performed by varying the components employed during ANN training, and the results are presented in Table 1. First one is the case only applying clip activation ($\phi_{Clip}$) from epoch 11 (I in Table 1), where the noticeable conversion loss is observed on all the parameter sets. This is mainly due to the two information losses after conversion, loss by input image encoding and loss by activation encoding. Second case is training ANN by applying TTFS activation ($\phi_{TTFS}$) to the input of the first layer, where the information loss of input images is small (I+II in Table 1). But, it still shows some conversion loss for more challenging datasets, like Tiny-ImageNet. This conversion loss further gets smaller in the third case, which is applying $\phi_{TTFS}$ to all the layers (I+II+III in Table 1). It is also noteworthy that the conversion loss reduction effect is more prominent when $T$ and $\tau$ are smaller. This is due to the small number of timesteps, which is similar to low bit width quantization.

Table 2 also shows the comparisons of the proposed CAT with T2FSNN [4]. The table presents the parameters used by kernel, the latency (timestep) of the network and the accuracies on CIFAR10, CIFAR100, and Tiny-ImageNet. The results in table 2 show that the proposed CAT achieves higher accuracy in all the cases. In terms of latency, the proposed CAT shows longer latency when same parameters are used, since T2FSNN utilizes 'Early Firing' technique to reduce latency. However, when $T$ and $\tau$ are reduced to smaller values, the proposed CAT shows better latency while showing better accuracies. One important thing to note is that, all the improvements presented in table 2 are achieved without any additional hardware cost. Moreover, the actual hardware cost can be reduced when using the proposed CAT since an identical kernel is used in the encoding and decoding processes of all the layers. The discussions on the hardware implementation cost will be presented in Section 4.

### 3.2 TTFS Coding for Logarithmic Computation

Although the proposed CAT efficiently reduces the conversion loss, complex multiplication operation to process each spike is a large burden from the hardware implementation point of view. In this work, to further reduce the complexity of spike processing, TTFS coding has been modified to process information in log-domain. Based on the observation that the spike time $t^l$ can be represented in log-domain, the multiplications between $w^l_{ij}$ and $\epsilon^l_{IN}$ in (7), can be replaced with much simpler operations, like

Table 1: The accuracies (conversion losses) of CAT

| Method | $T/\tau$ | CIFAR10 | CIFAR100 | Tiny-ImageNet |
|---|---|---|---|---|
| I | 48 / 8 | 92.32 (-1.33) | 67.93 (-4.55) | 58.75 (-2.28) |
| | 24 / 4 | 86.99 (-6.55) | 52.48 (-20.23) | 49.04 (-12.03) |
| | 12 / 2 | 62.78 (-30.69) | 15.07 (-57.52) | 17.19 (-43.84) |
| I+II | 48 / 8 | 92.85 (-0.23) | 70.62 (-1.06) | 59.31 (-1.61) |
| | 24 / 4 | 90.92 (-1.80) | 64.25 (-6.34) | 51.89 (-8.52) |
| | 12 / 2 | 78.21 (-12.98) | 33.93 (-33.27) | 21.18 (-37.88) |
| I+II+III | 48 / 8 | 93.18 (-0.02) | 71.72 (0.00) | 60.58 (-0.30) |
| | 24 / 4 | 92.45 (0.04) | 70.30 (-0.13) | 59.22 (-1.05) |
| | 12 / 2 | 90.77 (-0.05) | 66.00 (-0.56) | 54.99 (-3.90) |

I: Clip activation, II: TTFS activation applied to input of the first layer,
III: TTFS activation applied to all layers additions. To do this, we first rewrite the spike time $t^l$ using the kernel $\kappa^l$ that is specified in (9) as

$$t^l = \left\lceil \tau \log_2(u^l_i(t^l_{ref}-1)/\theta_0) \right\rceil + t^l_{ref}. \quad (14)$$

Table 2: Comparison with T2FSNN [4]

| | T2FSNN | This work | | |
|---|---|---|---|---|
| Base | e | e | 2 | 2 |
| $T$ | 80 | 80 | 48 | 24 |
| $\tau$ | 20 | 20 | 8 | 4 |
| Latency | 680 | 1360 | 816 | 408 |
| CIFAR10 | 91.43 | 93.36 | 93.18 | 92.45 |
| CIFAR100 | 68.79 | 72.14 | 71.72 | 70.30 |
| Tiny-ImageNet | - | 60.63 | 60.58 | 59.22 |

Like (8), (14) is not directly calculated during runtime in hardware. Then, $t^l$ can be expressed in log-domain, and it is used to remove the multiplication in (7). The logarithmic quantization is also used in ANN [13], [14], where feature maps and weights are quantized to power-of-2 logarithmic representation and multiplication is replaced with simple bit shift operation.

By adopting the post-training logarithmic quantization from [14], where the weights and the feature maps are quantized as follows:

$$\begin{aligned} w_q &= \text{sign}(w) a_w^{\widehat{w}_q} \\ \widehat{w}_q &= \text{clip}\bigl(\text{round}(\log_{a_w}|w| - FSR), 2 - 2^{bw-1}, 0\bigr) + FSR \end{aligned}, \quad (15)$$

where $w$ is the number to be quantized, $w_q$ is quantized version of $w$, full-scale range ($FSR$) is expressed as $\max(|w^l|)$, $bw$ is the quantization bit width, and $a_w$ is log-base. To replace the multiplication between $w^l_{ij}$ and $\epsilon^l_{IN}$ in (7) with a simple bit shift operation, the condition to be satisfied by the quantized operands is as follows [14]:

$$\log_2 a_w = 2^{-z_w}, \text{where } z_w \in \mathbb{Z}. \quad (16)$$

If (16) is satisfied, the multiplication in (7) can be rewritten as

$$\begin{aligned} p_q &= x_q w_q = \text{sign}(w_q) \, 2^{Int(\widehat{p_q})+Frac(\widehat{p_q})} \\ &= \text{sign}(w_q) \, 2^{Int(\widehat{p_q})} \, 2^{Frac(\widehat{p_q})} \\ &= \text{sign}(w_q) \, (LUT(Frac(\widehat{p_q})) << Int(\widehat{p_q})) \end{aligned}, \quad (17)$$

where $Int(x)$ and $Frac(x)$ are the integer part and the fractional part of input $x$, respectively. $LUT(k)$ is the content in the lookup table entry k and bit-wise operation $x << n$ is the left shift $x$ by $n$ bits.

To utilize (17) for the multiplication in (7), the proper selection of the parameters used in the kernel is needed to make $t^l$ meet (16). So, a constraint is added to $\tau$ as follow:

$$\log_2 \tau = 2^{z_\tau}, \text{ where } z_\tau \in \mathbb{Z}.  \quad (18)$$

Using logarithmic identity $n \log_a x = \log_{a^{1/n}} x$ and (18), it is observed that $t^l$ now satisfy (16), which means that TTFS coding can be implemented without multiplication. To select the kernel parameters and the quantization parameters for hardware implementation, the classification accuracies are simulated and the results are shown in Figure 4. According to the results shown in figure 4, $T = 24, \tau = 4, a_w = 2^{-1/2}$ and the weight bit width of 5 bits selected in our hardware implementation.

## 4 SUPPORTING SNN PROCESSOR DESIGN

### 4.1 Overall Architecture

Although the SNN obtained using the proposed methods can be deployed in general SNN processor architectures, SpinalFlow [7] is selected as the base of SNN processor in this work. SpinalFlow is selected because it employs an efficient dataflow to process SNNs using sparse temporal coding, such as TTFS coding in this work. The processing element (PE) in [7] is modified to support logarithmic computations, and spike encoder module is added to support the proposed TTFS coding.

The proposed SNN processor architecture is presented in figure 5, where the architecture can be divided into four parts, input generator, PE array, output processing, and control. The input generator consists of input buffer of 48KB and minfind unit to merge-sort input spikes. While the original SpinalFlow does not employ large input buffers, the input buffer of 48KB is used for reducing the number of DRAM accesses by increasing input reuse. The PE array has 128 PEs and four 90KB weight buffers, and it supports input gating to save energy when not all PEs are operating. The output processing in the proposed SNN architecture consists of post processing unit (PPU) and spike encoder, and it processes the output of PE array into spikes and saves the output spikes to output buffers before sending spike information to DRAM. The top control controls the overall architecture and direct memory access (DMA) engine manages data access to the off-chip DRAM. Input spikes are first processed by getting sorted in the input generator, and the sorted spikes are fed into the PE array to get accumulated into membrane voltage, which is the integration phase of TTFS coding used in this work. When the integration phase is over, the outputs of PEs are transferred to output processing to get encoded into output spikes (fire phase)

The detailed architecture of PE and spike encoding is presented in the right side of Figure 5. The spike encoder is designed to accommodate the proposed TTFS coding. It consists of membrane voltage (Vmem) buffer, comparators, threshold LUT, priority encoder and decoder. A priority encoder is used to handle the condition when multiple Vmems are larger than current threshold. The process of spike encoding is as follows: First, the process is initialized by moving Vmems from PPU to the Vmem buffer, where negative Vmems are set as zero since negative Vmems cannot produce spikes. Then, the encoding timestep starts from 1 and the dynamic threshold of current timestep are provided to the comparators by threshold LUT. For Vmems over the current threshold, comparator output is encoded to neuron ID by the priority encoder, and it is sent to the output buffer with current timestep. The output of priority encoder is fed back to the Vmem buffer to reset the Vmem that has just produced output spike. When all Vmems are smaller than current threshold, the encoding timestep increases by 1 and new threshold is provided to the comparators. This encoding process is repeated until all Vmems in the buffer are reset to zero or the end of last encoding timestep ($T$) is reached. After the end of the encoding process, output spikes in the output buffer are transferred to off-chip DRAM via DMA engine.

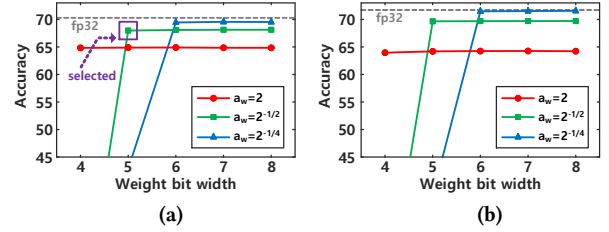

Figure 4: Accuracy vs. weight bit width on CIFAR100 with the kernel parameters of (a) $T = 24, \tau = 4$ (b) $T = 48, \tau = 8$

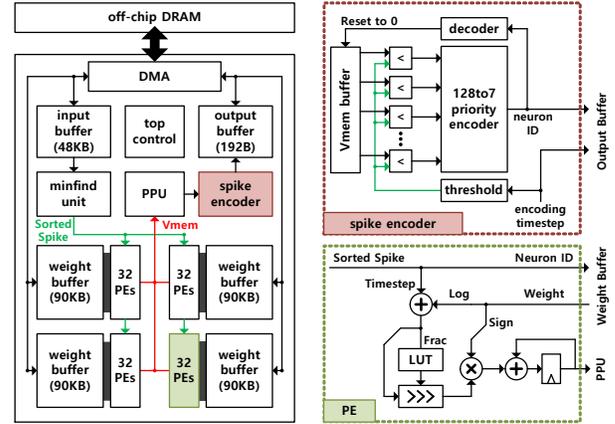

Figure 5: Overall SNN architecture with logarithmic PE and spike encoder module.

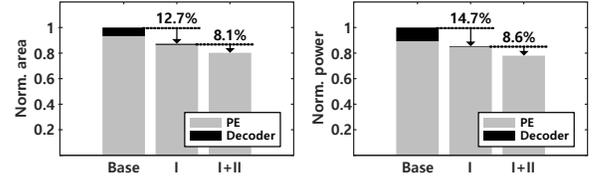

Figure 6: PE array area and power reduction of the proposed methods

## 5 EXPERIMENTAL RESULTS

The SNN processor to process the SNN model obtained from the proposed training methods has been designed and implemented using 28nm CMOS standard cell library. The design is first Verilog coded and it is synthesized with Synopsys Design Compiler using 28nm standard cell library. The power/energy consumption has been simulated using Synopsys PrimePower. DRAM access energy is calculated based on low energy HBM-like memory interface of 4 pJ/bit [15].

Figure 6 shows the area and power savings obtained from the proposed CAT (I) and the logarithmic computing TTFS coding (II). Here, the baseline architecture means the one that implement T2FSNN [4] using SpinalFlow. It also uses SRAM to decode spikes and employs multiplier (linear PE) to process the decoded spike. When only CAT is applied, the kernel parameters across all the layers are unified, thus spike decoder in PE array can be replaced into simple LUT from SRAM. So it achieves area and power savings of 12.7% and 14.7%, respectively. Additionally, when the logarithmic computing is employed, linear PE is changed to log PE, and it shows additional area and power savings of 8.1% and 8.6 %, respectively. Although the savings in area and power look small, it is important to note that those savings have been achieved while the accuracies of the SNN using the proposed training methods are improved compared to those of the conventional training methods.

Table 4 shows the comparisons with the previous works [10], [16]. As a popular ANN processor, TPU architecture is selected for comparison and it is redesigned to have 16×16 systolic array. Tianjic [10] is a state-of-the-art SNN processor that also process ANNs. In Table 4, the number of PEs in TPU means the number of MACs. The number of PE in [10] is obtained by (the number of cores) × (the number of MACs per core). Please note that the area of [10] is much larger than others as the number of PEs is larger and only on-chip memory is used. Comparing to ANN processor [16], the proposed SNN processor shows lower energy consumption and higher throughput thanks to the sparse-event computation of SNNs. In terms of accuracy, the proposed one shows relatively lower accuracies, but it can be improved if the quantization aware training is applied instead of post-training quantization to quantize the weights. Compared to the previous SNN processor [10], our design shows larger energy due to the utilization of off-chip DRAM and much deeper SNN model. In addition, since Tianjic [10] has significantly larger number of PEs (2496 versus 126), our SNN processor shows lower throughput. However, our design shows higher accuracy on CIFAR10, and it can also process CIFAR100 and Tiny-ImageNet dataset. To the best of our knowledge, our SNN processor is the first work to report the accuracy, throughput and energy of CIFAR100 and Tiny-ImageNet with hardware.

## 6 Conclusions

In this paper, we present the conversion aware training (CAT) method of utilizing multiple activation functions to simulate SNN during ANN training in order to reduce the ANN-to-SNN conversion loss and reduce the hardware burden at the same time. In addition, logarithmic information in spike is exploited to further reduce the hardware implementation costs. The hardware implementation results show that compared to the conventional SNN processor, our SNN processor is capable of processing deeper SNN with much improved accuracies. The proposed training methods can assist the design of hardware-friendly deep SNN with improved performance, and facilitate the use of SNNs to tackle more challenging tasks.

**Table 4: Comparison with previous ANN and SNN processors**

|  | This work | Tianjic [10] | TPU [16] (redesigned) |
|---|---|---|---|
| Type | SNN | SNN | ANN |
| Process | 28 nm | 28 nm | 28 nm |
| Voltage | 0.99 V | 0.85 V | 0.99 V |
| Area | 0.9102 mm$^2$ | 14.44 mm$^2$ | 1.4358 mm$^2$ |
| Memory Type | On-chip, Off-chip | On-chip | On-chip, Off-chip |
| Frequency | 250 MHz | 300 MHz | 250 MHz |
| Number of PEs | 128 | 2496 | 256 |
| Computational Throughput | 32 GSOP/s | 683.2 GSOP/s | 64 GMAC/s |
| Power | 67.3 mW | 950 mW | 100.1 mW |
| CIFAR10 | | | |
| Accuracy | 91.7 % | 89.5 % | 93.0 % |
| Energy per image | 486.7 uJ | 129 uJ | 978.5 uJ |
| Throughput | 327 fps | 46827 fps | 204 fps |
| CIFAR100 | | | |
| Accuracy | 67.9 % | - | 71.7 % |
| Energy per image | 503.6 uJ | - | 980.0 uJ |
| Throughput | 294 fps | - | 203 fps |
| Tiny-ImageNet | | | |
| Accuracy | 57.4 % | - | 61.4 % |
| Energy per image | 1426 uJ | - | 2759 uJ |
| Throughput | 63 fps | - | 51 fps |